\documentclass[11pt]{article}

\usepackage[final]{acl}

\usepackage{times}
\usepackage{latexsym}
\usepackage{booktabs}
\usepackage{multirow}
\usepackage{graphicx}
\usepackage{amsmath}
\usepackage{amssymb}

\usepackage[T1]{fontenc}
\usepackage[utf8]{inputenc}

\usepackage{microtype}

\usepackage{inconsolata}

\usepackage{graphicx}

\ifdefined\iflatexml
\else
    \newif\iflatexml
\fi

\title{Less is More: Label-Guided Efficient Summarization of Procedural and Instructional Videos}
\iflatexml
\author{
{\normalfont\mdseries Shreya Rajpal}\thanks{Work done during an internship at the University of Tübingen.} \\
Vellore Institute of Technology \\
\texttt{shreyarajpal6@gmail.com}
\and
{\normalfont\mdseries Michal Golovanevsky} \\
Brown University \\
\texttt{michal\_golovanevsky@brown.edu}
\and
{\normalfont\mdseries Carsten Eickhoff}\thanks{Corresponding author.} \\
University of T\"{u}bingen \\
\texttt{carsten.eickhoff@uni-tuebingen.de}
}
\else
\author{
Shreya Rajpal$^{1}$\thanks{Work done during an internship at the University of Tübingen.} \quad
Michal Golovanevsky$^{2}$ \quad
Carsten Eickhoff$^{3}$\thanks{Corresponding author.} \\
$^{1}$Vellore Institute of Technology \quad
$^{2}$Brown University \quad
$^{3}$University of Tübingen \\
\texttt{shreyarajpal6@gmail.com} \quad
\texttt{michal\_golovanevsky@brown.edu} \\
\texttt{carsten.eickhoff@uni-tuebingen.de}
}
\fi

\begin{document}
\maketitle
\begin{abstract}
Video summarization transforms long videos into concise representations that are easier to document and analyze, especially in high-stakes domains such as surgical training. However, long-form videos often require dense frame processing or supervised training pipelines, which can be computationally expensive and may still miss important procedural content. We present PRISM: Procedural Representation via Integrated Semantic and Multimodal Analysis, a zero-shot and training-free framework for frame-efficient procedural video summarization. PRISM selects fewer than 5\% of video frames while retaining over 84\% semantic content and improving over baselines by up to 7.5\%. Rather than exhaustively processing frames, PRISM uses lightweight visual filtering and dynamically generated procedural labels as semantic anchors to select meaningful frame-label groups. This selective inference design preserves key actions, transitions, and contextual details while reducing the number of visual inputs passed to downstream vision-language captioning stages. We evaluate PRISM on YouCook2 and ActivityNet Captions, with additional studies on keyframe selection benchmarks and surgical video datasets. Across procedural and domain-specific video tasks, PRISM achieves strong semantic alignment and accuracy, suggesting that efficient multimodal video understanding can be achieved by grounding generation in dynamically generated semantic anchors.
\end{abstract}

\section{Introduction}

Video summarization plays a significant role in making long-form video content easier to analyze, especially in domains such as robotics \citep{robot}, education \citep{edu}, and surgical training \citep{surg}. However, long-form instructional and surgical videos often exceed 40 minutes \citep{autolap, lin2025unleashing}, making it challenging to process a large number of frames. Therefore, effective summarization must reduce visual processing while preserving important actions, transitions, and procedural context.

A common approach to video summarization is keyframe selection, which extracts informative frames or segments that best represent the content of a video. In high-stakes domains such as surgical video summarization, summaries must do more than shorten the video; they must preserve the order of actions, procedural intent, and contextually important details. This creates a central challenge for long-form procedural summarization, as systems must be computationally efficient while still retaining the semantic and temporal structure needed for accurate summaries. \footnote{Code is available at \url{https://github.com/Shreyarajpal12/PRISM-code}.}

Existing methods often emphasize either summary quality or efficiency alone. Lightweight approaches based on visual cues, motion, similarity, or clustering reduce redundancy but may miss fine-grained procedural transitions \citep{Histogram, motion, surg2, SumMe, tvsum}. In contrast, deep sequence models, adversarial learning, attention mechanisms, pretrained embeddings, and vision-language models improve semantic relevance and temporal coherence, but often rely on supervised frame-level scorers, video-level captions, predefined queries, rule-based selection, or task-specific training \citep{deepseq, sumgan, pretraining, radford, blip, clipit, cap2sum, lmske, tacsum}. This motivates a summarization framework that can reduce the number of visual inputs while preserving procedural meaning without requiring supervised annotations or fixed queries. 

In this work, we present PRISM: Procedural Representation via Integrated Semantic and Multimodal Analysis, a zero-shot and training-free framework for frame-efficient video summarization. PRISM is designed to maintain both efficiency and semantic grounding. Rather than relying on dense frame processing, PRISM first uses lightweight visual filtering to identify candidate frames with meaningful transitions. It then dynamically generates procedural labels from the video itself, validates them as semantic anchors, and aligns them with relevant frames using vision-language similarity. The selected frame-label groups are temporally aggregated and summarized into a coherent procedural narrative. This allows PRISM to reduce redundant visual inputs while preserving key actions, transitions, and contextual details.

\begin{figure*}[t]
    \centering
    \includegraphics[width=0.95\textwidth]{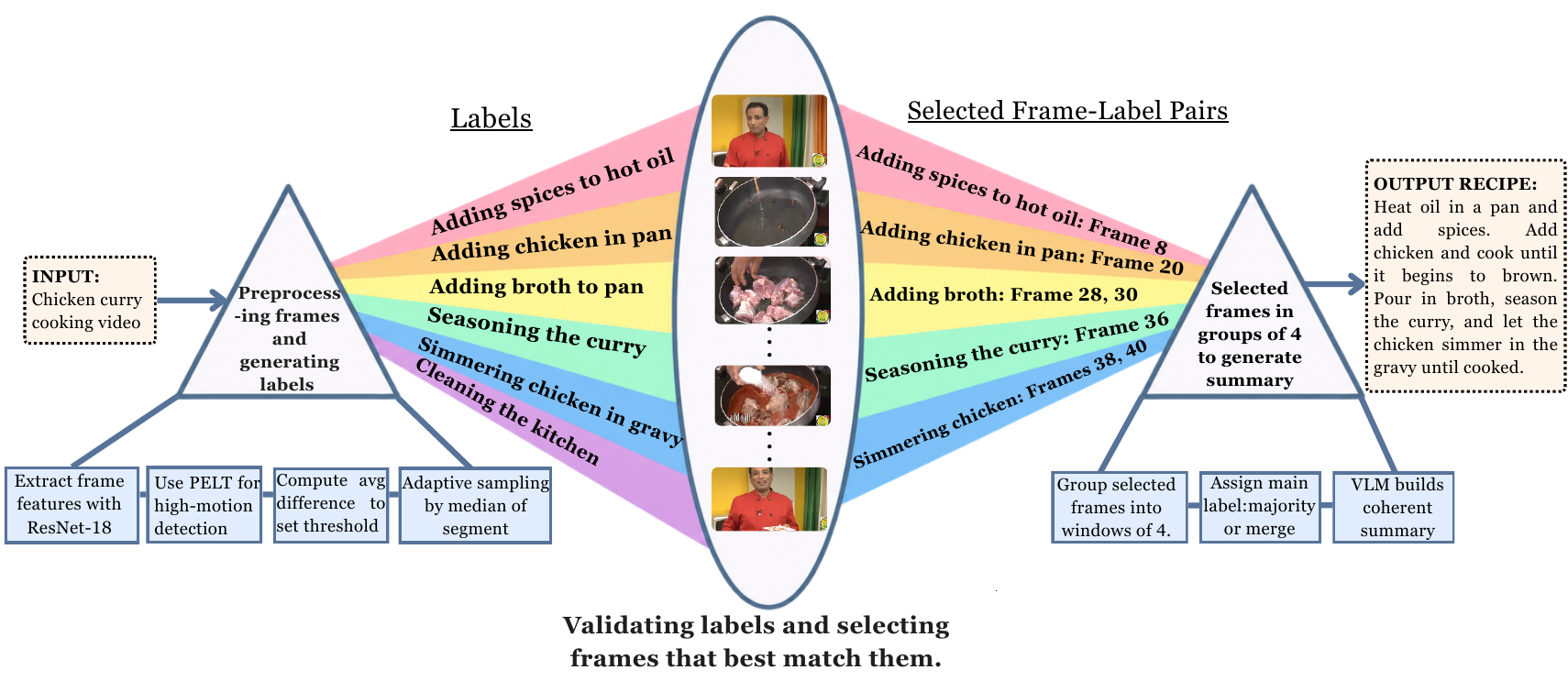}
    \caption{PRISM pipeline: the input video is semantically decomposed, filtered, and recombined into a coherent summary.}
    \label{fig:prism_pipeline}
\end{figure*}

The framework draws inspiration from the behavior of light through prisms. The left prism decomposes raw video into semantically meaningful elements using frame embeddings and adaptive sampling. The center filtering stage validates labels and selects matching frames, acting as a lens that sharpens and aligns relevant content. The right prism recombines the selected frame-label pairs into a focused semantic summary using language models. Like light passing through two prisms and a filter, PRISM decomposes procedural video content, filters it through semantic anchors, and reassembles it into a coherent, context-rich narrative.

We evaluate PRISM on YouCook2 \citep{youcook2} and ActivityNet Captions \citep{act} for video-level captioning and summarization. PRISM selects fewer than 5\% of video frames while preserving over 80\% of semantic content, measured using BERTScore. It also improves over existing baselines on video-level captioning tasks, measured using METEOR \citep{meteor}. We further study PRISM in medical video summarization, where reducing visual processing while preserving procedural structure is especially important. We also do an additional keyframe selection study on TVSum \citep{tvsum} and SumMe \citep{SumMe} and report results in Appendix F.

Our contributions are:
\begin{itemize}
    \item We introduce PRISM, a zero-shot and training-free framework for frame-efficient procedural video summarization that does not rely on supervised frame-level labels, fixed user queries, or dataset-provided annotations during inference.

    \item We propose a label-guided selective inference pipeline that combines lightweight visual filtering with dynamically generated procedural labels. These labels serve as semantic anchors for selecting meaningful frame-label groups, reducing dense visual processing while preserving procedural structure.

    \item We show that dynamically generated labels improve the grounding of vision-language summarization by guiding frame selection through vision encoder similarity, enabling summaries to be constructed from semantically selected evidence rather than densely processed frames.

    \item We evaluate PRISM across instructional, activity, keyframe selection, and surgical video settings. Results show that PRISM selects fewer than 5\% of frames while maintaining strong semantic alignment and precision, demonstrating efficient multimodal video understanding without supervised training or task-specific fine-tuning.
\end{itemize}

\section{Related Works}

Video summarization transforms long-form videos into compact representations, including static keyframes \citep{static}, dynamic skim videos \citep{dynamic}, and textual summaries. Early methods estimated keyframe importance using visual or statistical cues such as color-histogram clustering \citep{vsum}, motion and optical-flow activity \citep{motion, opticalmotion}, and entropy-based selection \citep{entropy}. Later approaches formulated summarization as visual-importance prediction through LSTM-based frame scorers \citep{lstm}, adversarial and reinforcement-learning frameworks \citep{sumgan, zhou}, and attention-based models such as VASNet and CSTA for temporal or spatio-temporal modeling \citep{vasnet, csta}. These methods improve frame selection, but many treat summarization primarily as a visual importance scoring problem and depend on human annotations.

Recent methods incorporate language, audio, captions, and vision-language models to improve semantic relevance and narrative flow. Query-focused and personalized methods align summaries with user queries or preferences \citep{sharghi, Zhangquery, person, interest}. 
Caption-based methods use language as a semantic bridge, FrameRank ranks captioned frames \citep{framerank}, CLIP-It aligns frames to video-level dense captions using CLIP \citep{radford, clipit}, Cap2Sum uses dense captions as supervision \citep{cap2sum}, and LMSKE applies shot segmentation and CLIP-feature clustering for sequential keyframe extraction \citep{lmske}.
LLMVS uniformly samples frames, captions them with a multimodal LLM, and trains an LLM-based visual importance scorer using human annotations \citep{llms}, while ZeroTA performs dense video captioning in a zero-shot setting by optimizing text and moment representations at inference time \citep{zerota}. 
Although these methods improve semantic grounding, many rely on external queries, global or dense captions, user preferences, human annotations, or supervised training. In contrast, PRISM is training-free, reduces the candidate frame set before VLM captioning, and uses dynamically generated procedural labels as semantic anchors rather than ranking uniformly sampled frames.

Beyond summarization accuracy, efficiency and computational sustainability remain central challenges in long-form video understanding.
Recent efficient video-language methods reduce visual processing through query-conditioned frame selection, token compression, or saliency-aware retrieval.
Multimodal large language model-based frame selection and Frame-Voyager select frames conditioned on a question or task query \citep{yu2025framevoyager}; PruneVid and DyCoke reduce redundant visual tokens or KV-cache computation inside video-language models \citep{huang2025prunevid, tao2025dycoke}; and Sali4Vid improves dense video captioning through saliency-aware frame reweighting derived from timestamp annotations and semantic-segment-based caption retrieval\citep{sali4vid}. These approaches demonstrate the value of efficient visual processing, but they either require an external task signal, reduce computation after visual inputs have entered the model, or rely on annotation-derived supervision.

PRISM combines frame-level efficiency with bottom-up semantic grounding. It requires no external query, user profile, global caption, dense caption supervision, human annotations, event timestamp, or dataset-provided annotation. Instead, PRISM generates procedural labels from sampled frames, validates them against the video's own context, and uses them as semantic anchors for selecting frame-label pairs through vision-language similarity. Because this selection occurs before downstream summarization, PRISM reduces the number of visual inputs passed to downstream VLM stages to fewer than 5\% of the original frames. 

\section{Method}
Our proposed framework, PRISM, is a multistage pipeline that enables efficient and context-aware keyframe selection, followed by semantically grounded video summarization. This pipeline is designed to ensure that the summary covers key visuals, preserves contextual meaning and procedural order, and remains efficient across domains.
\subsection{Dataset and Experimentation}
Our pipeline targets video summarization, with a focus on video-level dense captioning. We use YouCook2~\citep{youcook2}, a dataset of over 2,000 untrimmed instructional cooking videos spanning 89 recipes, each annotated with temporal segments and stepwise imperative captions. We also use the ActivityNet Captions \citep{act} dataset, which contains temporally localized, natural language descriptions for untrimmed videos, making it a benchmark for dense video captioning and summarization tasks. We evaluate summarization performance using BLEU, ROUGE-L, METEOR, and BERTScore~\citep{bleu, rouge,Bertscore,meteor} with a focus on capturing semantic alignment, procedural coherence, and linguistic quality. Lastly, we briefly discuss the system’s application in the medical domain.

\subsection{Stage 1: Frame Embedding \& Adaptive Sampling}

\begin{figure*}[t]
    \centering
    \includegraphics[width=0.65\textwidth]{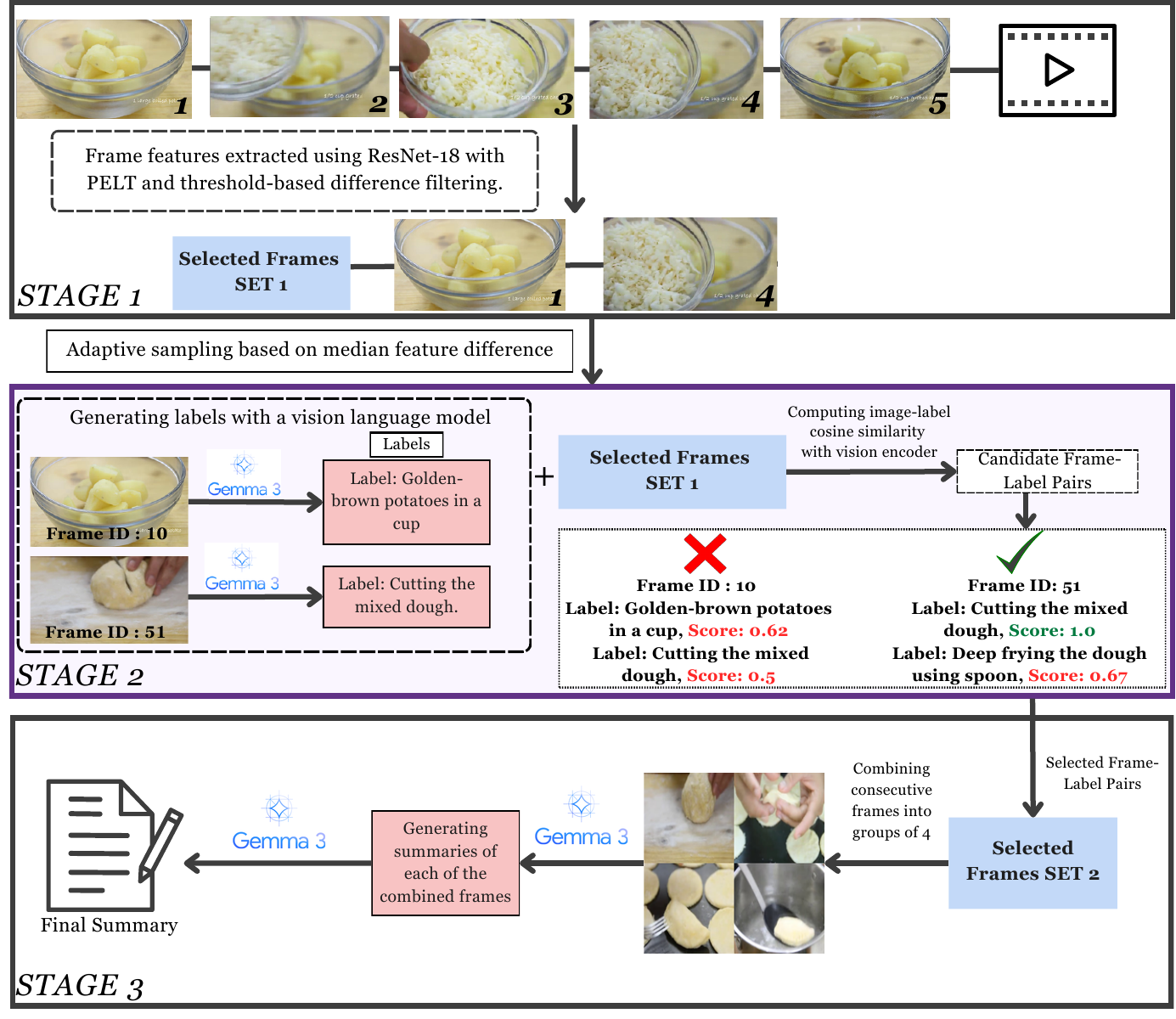}
    \caption{A complete example of the workflow.}
    \label{fig:complete}
\end{figure*}
We first convert the raw video into a manageable set of candidate frames. Frames are sampled at a fixed rate (e.g., 1 fps) and passed through ResNet-18 to extract 512-dimensional embeddings $f_i \in \mathbb{R}^{512}$ that capture high-level spatial features~\citep{resnet}.
Next, we detect statistically significant visual transitions via the Pruned Exact Linear Time (PELT) algorithm \citep{pelt}. PELT identifies a set of change indices $\{t_k\}$ such that embedding differences before and after each $t_k$ exceed a model-cost penalty, yielding an initial subset $F_{\text{PELT}} \subset \{f_i\}$.

To further adapt sampling density to visual variability, we partition the video into segments between change points and compute, for each segment $s$ (here, we set $s$=10), the median Euclidean distance $\text{med}_s$ among its frame embeddings, and $\delta$ is the median feature-difference threshold. Segments with $\text{med}_s < \delta$ retain one representative frame; those with $\text{med}_s \geq \delta$ retain two. In our video experiments, we set $\delta = 0.30$. The resulting visually filtered and adaptively sampled frames form Selected Frames Set 1, which is passed to the semantic anchoring stage as shown in Figure~\ref{fig:complete}. The nuances of different hyperparameter settings are discussed in the Appendix.

\subsection{Stage 2: Label Generation and Semantic Anchoring}

Following adaptive frame sampling, the induced frame set $F = \{f_i\}_{i=1}^n$ is processed to generate semantic anchors that represent high-level procedural steps. This stage involves three subprocesses: caption generation, semantic label validation, and label to frame association.

We begin by feeding each sampled frame $f_i$ into a vision-language model. Depending on the domain and setup, models such as Gemma3 \citep{gemma3}, LLaMA 3.2 Vision \citep{llama3}, MiniCPM \citep{minicpmv}, or GPT4 \citep{gpt} are employed to generate per-frame captions that describe the localized content.
These captions are generalized into anchor labels. For example, the caption \emph{"The person in a white apron and clear gloves is carefully sprinkling a large amount of shredded cabbage into a large stainless steel bowl..."} is mapped to the label \emph{"Sprinkling shredded cabbage for kimchi"}, which may span multiple frames.

To ensure label quality, we introduce a label validation step using an LLM (in this case GPT-4). The model filters out labels that are overly vague (e.g., ``Cooking instructions being given``) or contextually insignificant (e.g., ``cleaning utensils`` is insignificant to the recipe)  and retains only those labels that represent discrete procedural events. Stage 2 and Stage 3 can reduce noisy or hallucinated labels by checking whether they are coherent with the broader procedural context and visually aligned with selected frames. For example, in a recipe about ground beef, a hallucinated label referring to chicken may be removed if it is inconsistent with the remaining generated labels or weakly aligned with the visual evidence. Such frames are discarded whenever possible, as they do not align with the procedural context. This step also eliminates visually insignificant frames that may be missed by ResNet-based analysis, such as black screens, frames dominated by light flicker or glare, and transition frames without identifiable objects, which do not produce meaningful captions. As illustrated in Figure~\ref{fig:complete}, this process filters the initial selected frames and labels into high-confidence frame-label pairs, which form Selected Frames Set 2 for downstream summarization.

\begin{figure}[t]
  \centering
 \includegraphics[width=0.4\textwidth]{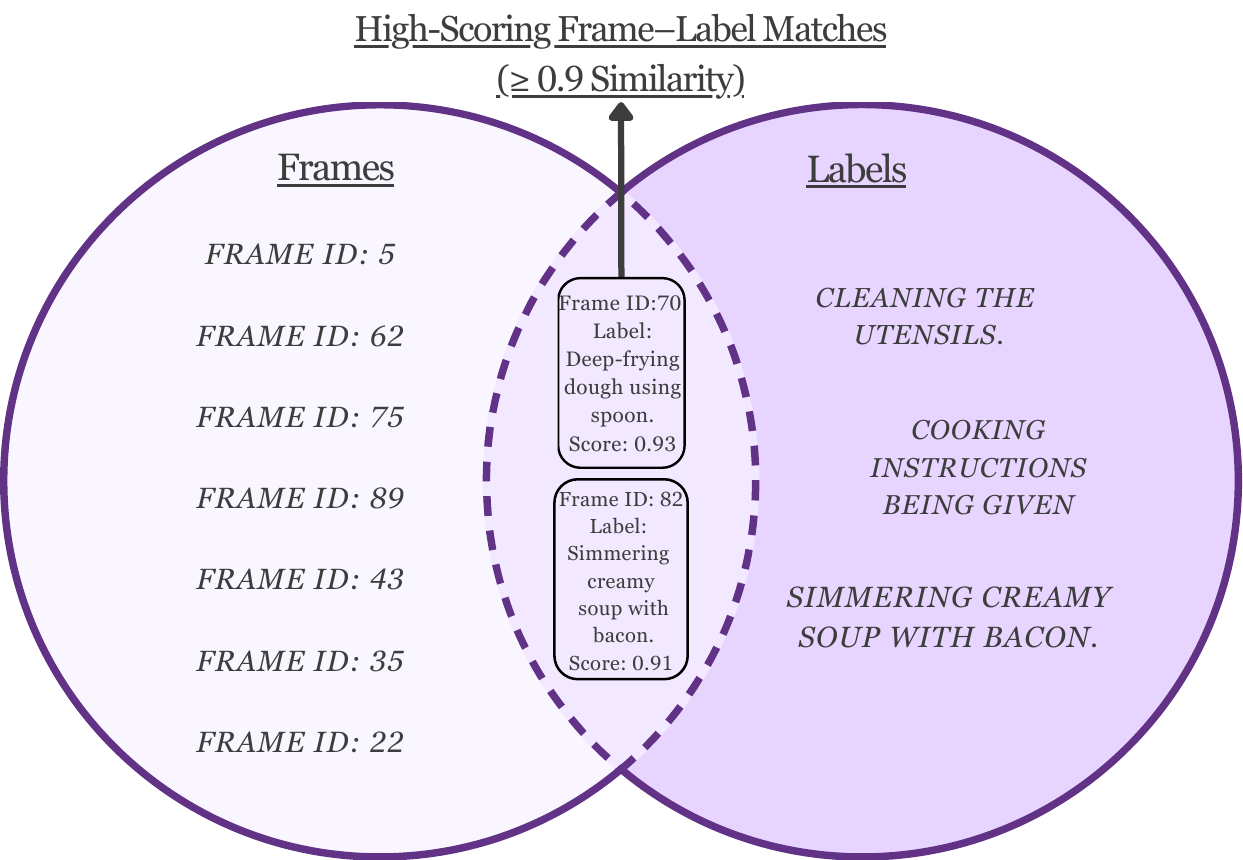}
  \caption{A Venn diagram illustrating overlap between frame embeddings and semantic labels where cosine similarity exceeds 0.9. Highlighted labels reflect procedural relevance across frames.}
  \label{fig:venn}
\end{figure}

The Venn diagram in Figure \ref{fig:venn} shows frame-label pairs at a similarity threshold of 0.9, highlighting only high-confidence matches (e.g., "Deep frying dough" at 0.93) for summarization. Frames and labels outside this intersection are excluded from the summary due to weak or irrelevant alignment.
Next, we perform label to frame association using vision encoders such as CLIP by ~\citep{radford}, BLIP by \citep{blip}, or BioMedCLIP by \citep{biomedclip}. Each label $\ell_j$ and frame $f_i$ is embedded into a shared space, and cosine similarity is computed:

\[
S(f_j, \ell_i) = \frac{\langle E(f_j), E(\ell_i) \rangle}{\|E(f_j)\| \cdot \|E(\ell_i)\|}
\]
\begin{itemize}
    \item $\mathcal{F} = \{f_1, f_2, \ldots, f_N\}$: sampled frames
    \item $\mathcal{L} = \{\ell_1, \ell_2, \ldots, \ell_M\}$: validated semantic labels
    \item $\text{sim}(f_i, \ell_j)$: cosine similarity between $f_i$ and $\ell_j$
    \item $\tau$: similarity threshold (set to $0.9$)
\end{itemize}

We assign labels to frames using:
\begin{equation}
A(f_i)=
\begin{cases}
\ell_i^{*}, & \text{if } \max_{\ell \in \mathcal{L}} \operatorname{sim}(f_i,\ell) \geq \tau,\\
\emptyset, & \text{otherwise,}
\end{cases}
\end{equation}
Here, $\ell_j^{*}$ denotes the best-matching semantic label. Frames with no label exceeding the threshold $\tau$ are discarded. This ensures semantic precision and relevance.

Notably, a frame may match multiple labels (e.g., 0.98 for $\ell_a$ and 0.95 for $\ell_b$), reflecting overlapping steps in procedures. These frames are retained and prioritized during summarization.

\subsection{Stage 3: Temporal Aggregation and Summary Construction}

We now organize the selected frames into temporally ordered groups of four, forming windows:

\[
W_i = \{f^*_{i-1}, f^*_i, f^*_{i+1}, f^*_{i+2}\}
\]
Here, $f_i^{*}$ denotes the filtered frame obtained after Stage 2.
Each window is passed into a vision-language summarizer (e.g., GPT4V, Gemini Vision, MiniCPM-V) to generate localized summaries.
Each 4-frame group is assigned a main label based on frame-wise captions. If all labels differ, the main label combines all four. If two labels occur twice (2:2), both are used. Otherwise, the majority label is assigned (e.g., 3:1).

These summaries are then stitched together using a large language model (e.g., GPT-4, LLaMA 3.1) following a layered, tree-like approach. Summaries are grouped based on the model’s maximum context size and processed at the leaf level. The resulting outputs are recursively merged at higher levels, enabling the generation of a final, cohesive summary that integrates information from all segments in a structured manner. This step isolates frames and reduces hallucinations or inconsistencies by anchoring each part within the broader narrative.

Figure \ref{fig:complete} represents the overview of our three-stage video summarization framework. Stage 1 extracts frame embeddings using ResNet-18 by \citep{resnet, resnet1}, with adaptive sampling based on thresholded feature difference and PELT segmentation. Stage 2 generates semantic labels using a vision-language model (e.g., Gemma3), followed by label-to-frame similarity scoring using a vision encoder. Only frames with cosine similarity exceeding a threshold (e.g., 0.9) are retained. 

Stage 3 combines temporally adjacent filtered frames into groups of four to retain local visual context and generate a summary caption for each group. This helps in progressively constructing the final output narrative. Since the frames have already been filtered and matched with labels, each group contains representative moments rather than consecutive frames. We use four frames as a practical window size, since smaller groups may split a single procedural action, while larger groups may mix different steps and increase inference cost. 
This multi-stage process yields a coherent, semantically grounded video summary suitable for high-stakes domains like surgery, technical instruction, or training.

% Figure~\ref{fig:complete} illustrates the complete workflow of the proposed framework. In Stage 1, ResNet-18 embeddings, PELT-based change-point detection, and threshold-based filtering remove redundant or visually similar frames, producing Selected Frames Set 1. Adaptive sampling then improves coverage by adding frames that capture meaningful visual transitions. In Stage 2, semantic labels generated by a vision-language model are aligned with the selected frames using image-label cosine similarity from a vision encoder. This filters weak frame-label matches and retains high-confidence pairs; for example, Frame ID 51 is retained with the label ``Cutting the mixed dough'' because it achieves a similarity score of 1.0. These retained frame-label pairs form Selected Frames Set 2, which serves as the final visual evidence for summarization. In Stage 3, the selected frames are grouped into windows of four, summarized locally by the vision-language model, and combined to construct the final video summary. Overall, the workflow shows how PRISM progressively reduces raw video frames into semantically grounded frame-label groups before generating the final narrative.

\section{Results}
We evaluate our method, PRISM, across dense video captioning tasks, benchmarking it against prior video summarization models using standard datasets and evaluation metrics. The results are reported using the MiniCPM-V model. Scores are scaled to 0–100 for all metrics.

\subsection{Video Level Summarization}
We first evaluate summarization quality on the YouCook2 dataset and the ActivityNet Captions dataset. Table~\ref{tab:youcook2_rouge_meteor} compares PRISM against multimodal and vision-only baselines using ROUGE-L, METEOR, and BERTScore. While existing works like UniVL and COOT \citep{meltr} \citep{coot} focus on leveraging paired video-text pretraining, PRISM achieves strong semantic alignment, particularly outperforming all models on METEOR Score. These results highlight the strength of our summarization approach even with significantly fewer processed frames.

The evaluation uses a mix of traditional and semantic-based metrics. While BLEU and ROUGE are commonly used in video summarization, they fall short for our approach, which prioritizes generating long, detailed summaries over short, high-level captions as given in the dataset. 
\begin{table*}[t]
\centering
\scriptsize
\setlength{\tabcolsep}{3pt}
\renewcommand{\arraystretch}{0.95}

\begin{tabular}{lcccc}
\toprule
\textbf{Model} & \textbf{Modality} & \textbf{ROUGE-L} & \textbf{METEOR} & \textbf{BERT} \\
\midrule
ZeroTA~\citep{zerota}              & V    & --    & 2.10  & -- \\
Streaming GIT~\citep{stream}       & V    & --    & 3.60  & -- \\
PDVC (CLIP)~\citep{pdvc,v2s}       & V    & --    & 5.70  & -- \\
CM$^2$~\citep{cm}                  & V    & --    & 6.08  & -- \\
E$^2$DVC~\citep{e2dvc}             & V    & --    & 6.11  & -- \\
CACMI~\citep{cacmi}                & V    & --    & 6.21  & -- \\
PR-DETR~\citep{prdetr}             & V+T  & --    & 6.48  & -- \\
Streaming Vid2Seq~\citep{stream}   & V    & --    & 7.10  & -- \\
Vid2Seq~\citep{v2s}                & V    & --    & 9.30  & -- \\
VideoBERT~\citep{vbert}            & V    & 27.14 & 10.81 & -- \\
EMT~\citep{emt}                    & V    & 27.44 & 11.55 & -- \\
HiCM$^2$~\citep{hicm2}             & V    & --    & 12.80 & -- \\
CBT~\citep{cbt}                    & V    & 30.44 & 12.97 & -- \\
ActBERT~\citep{abert}              & V    & 30.56 & 13.30 & -- \\
Sali4Vid~\citep{sali4vid}          & V    & --    & 13.54 & -- \\
MCCL~\citep{mccl}                  & V    & --    & 14.69 & -- \\
UniVL~\citep{univl}                & V    & 40.09 & 17.57 & -- \\
AT+Video~\citep{atv}               & V+T  & 36.65 & 17.77 & -- \\
DPC~\citep{dpc}                    & V+T  & --    & 18.08 & -- \\
UniVL + MELTR~\citep{meltr}        & V    & 41.28 & 18.19 & -- \\
COOT~\citep{coot}                  & V    & 37.94 & 19.85 & -- \\
UniVL~\citep{univl}                & V+T  & 46.52 & 22.35 & -- \\
UniVL + MELTR~\citep{meltr}        & V+T  & 47.04 & 22.56 & -- \\
\midrule
\textbf{PRISM (Ours)}              & V    & 16.49 & \textbf{23.66} & \textbf{82.78} \\
\textbf{PRISM (Ours)}              & V+T  & 23.58 & \textbf{30.08} & \textbf{84.34} \\
\bottomrule
\end{tabular}
\caption{Performance comparison on YouCook2 using ROUGE-L, METEOR, and BERTScore.}
\label{tab:youcook2_rouge_meteor}

\end{table*}
For instance, the sentence ``A young woman is seen standing in a room and leads into her dancing'' is the ground-truth caption for a video segment in ActivityNet Captions. In contrast, our generated summary captures more detailed dance movements: ``The video begins with a ballet dancer's movement in the en croisé position, emphasizing the precision, balance, and control required to execute the transition smoothly.'' BLEU and ROUGE mainly reward surface overlap through n-grams and longest common subsequences, so they can underrepresent semantically accurate but more detailed summaries. We therefore emphasize METEOR as the primary metric for comparison.
\begin{table*}
\centering
\scriptsize
\setlength{\tabcolsep}{4pt}
\renewcommand{\arraystretch}{0.95}

\begin{tabular}{lcc}
\toprule
\textbf{Model} & \textbf{BLEU-4} & \textbf{METEOR} \\
\midrule
ZeroTA~\citep{zerota}          & --    & 2.7 \\
Vid2Seq~\citep{v2s} & -- & 8.50 \\
CM$^2$~\citep{cm} & 2.38 & 8.55 \\ 
E$^2$DVC~\citep{e2dvc}    & 2.43 & 8.57 \\
PR-DETR~\citep{prdetr} & 2.58 & 8.72 \\ 
TSP~\citep{tsp} & 2.02 & 8.75 \\ 
Streaming GIT~\citep{stream} & -- & 9.00 \\
Streaming Vid2Seq~\citep{stream} & -- & 10.00 \\
GVL~\citep{gvl} & 1.11 & 10.03 \\
BMT~\citep{bmt} & 1.99 & 10.90 \\
PDVC (TSN)~\citep{pdvc} & 3.07 & 11.27 \\ 
iPerceive~\citep{ip} & 2.98 & 12.27 \\
MCCL~\citep{mccl} & 3.89 & 12.52 \\ 
ADV-INF + Global~\citep{adv} & 9.45 & 16.36 \\
\midrule
\textbf{PRISM (Ours)} & 1.02 & \textbf{20.04} \\
\bottomrule
\end{tabular}
\caption{Performance comparison on ActivityNet Captions using BLEU-4 and METEOR.}
\label{tab:activitynet_bleu_meteor}

\end{table*}
V and V+T in Table~\ref{tab:youcook2_rouge_meteor} refer to PRISM's vision-only and vision+text modalities, respectively, where text denotes the audio transcript. PRISM improves over the YouCook2 baseline by 7.5\% in METEOR, as shown in Table~\ref{tab:youcook2_rouge_meteor}. Table~\ref{tab:activitynet_bleu_meteor} compares state-of-the-art models on the ActivityNet Captions validation set. PRISM achieves the highest METEOR score of 20.04, surpassing the strongest baseline by 3.68\% relatively.
\subsection{Key frame selection}

To capture deeper semantic alignment, we also adopt LLM-based evaluation in ablation studies and case studies, where a language model serves as a judge to assess summary quality based on factual accuracy, detail, specificity, completeness and repetition. Each of these aspects is scored on a scale from 1 to 5. We then compute a weighted average, applying double weight to factual accuracy, and normalize the result to a score between 0 and 1. This forms our normalized LLM Judge Score.

\subsection{Case study: Medical Datasets Video Summarization}
\begin{table}[t]
\centering
\scriptsize
\setlength{\tabcolsep}{2pt}
\renewcommand{\arraystretch}{0.95}

\begin{tabular}{llccc}
\toprule
\textbf{Dataset} & \textbf{Model} & \textbf{BERT} & \textbf{Sem-nCG} & \textbf{Phase Cov.} \\
\midrule
\multirow{3}{*}{Cholec80}
 & LLaMA 3.2-Vision & 87.04 & 76.67 & 79.01 \\
 & Gemma 3          & \textbf{87.33} & 77.08 & 77.91 \\
 & MiniCPM          & 86.93 & \textbf{81.11} & \textbf{80.83} \\
\midrule
\multirow{3}{*}{PIT-VIS}
 & LLaMA 3.2-Vision & \textbf{85.05} & \textbf{100.00} & 74.36 \\
 & Gemma 3          & 84.68 & \textbf{100.00} & 79.13 \\
 & MiniCPM          & 84.75 & 97.22 & \textbf{81.95} \\
\bottomrule
\end{tabular}
\caption{Medical case study results on Cholec80 and PIT-VIS. MiniCPM achieves strong semantic and procedural summarization scores and is used as part of the pipeline. Phase coverage is computed using LLM-based matching.}
\label{tab:medical_case_study}

\end{table}
We explore the applicability of our summarization pipeline in medical settings through a case study on two surgical datasets: Cholec80 and PIT-VIS. Our pipeline uses MiniCPM to generate summaries, which are then evaluated using BERTScore, Sem-nCG, and LLM as a judge for phase level coverage \citep{semncg}.

To assess procedural completeness, we extracted 7 standard surgical phases in Cholec80 and 12 annotated phases in PIT-VIS (excluding pre/post phases with insufficient volume) and checked their presence within the generated summaries using a large language model. On average, MiniCPM captured 5.6/7 phases in Cholec80 and 9.24/12 in PIT-VIS, indicating strong phase-level alignment without direct supervision.

This experiment was conducted to investigate the feasibility of training-free textual summarization pipelines that could combine multiple surgical subtasks e.g., frame-level labeling, phase annotation, and summary generation, into a unified algorithm. While we do not position these results as clinically validated, they point to a promising direction for automated surgical video understanding. We invite future research and clinical collaborations to rigorously evaluate and refine such pipelines for real-world deployment.
\section{Ablation Studies}

\begin{table*}[t]
\centering
\scriptsize
\setlength{\tabcolsep}{1.8pt}
\renewcommand{\arraystretch}{0.9}

\begin{tabular}{llcccccccc}
\toprule
\textbf{Data} & \textbf{Setting} & \textbf{BLEU} & \textbf{R-L} & \textbf{BERT} & \textbf{MET} & \textbf{Judge} & \textbf{Frm} & \textbf{Time} & \textbf{Tok} \\
\midrule
\multirow{4}{*}{YC}
 & Video-Only    & 1.82 & 18.40 & 82.41 & 19.83 & 38.64 & 22.25  & 71.78   & 3{,}560 \\
 & No Stage 2    & 3.00 & 28.11 & 84.12 & 31.83 & 79.26 & 60.07  & 73.09   & 9{,}613 \\
 & No Stage 1    & 2.80 & 29.87 & 84.59 & 32.66 & 78.49 & 22.23  & 236.46  & 3{,}557 \\
 & No Proc.      & 2.70 & 28.03 & 84.13 & 31.83 & 80.82 & 322.91 & 1625.40 & 206{,}662 \\
\midrule
\multirow{4}{*}{AN}
 & Video-Only    & 0.97 & 9.04 & 80.98 & 17.29 & 37.54 & 18.39  & 40.37  & 2{,}942 \\
 & No Stage 2    & 0.99 & 8.77 & 81.19 & 17.59 & 43.33 & 20.04  & 28.52  & 3{,}206 \\
 & No Stage 1    & 0.97 & 8.68 & 81.29 & 17.47 & 43.57 & 17.39  & 88.69  & 2{,}782 \\
 & No Proc.      & 0.91 & 7.87 & 80.56 & 16.09 & 43.75 & 111.88 & 130.06 & 71{,}603 \\
\bottomrule
\end{tabular}
\caption{Ablation results on YouCook2 (YC) and ActivityNet Captions 
(AN) across pipeline settings. \textbf{BLEU}, \textbf{R-L} (ROUGE-L), 
\textbf{BERT} (BERTScore), \textbf{MET} (METEOR), and \textbf{Judge} 
(LLM-Judge) are quality metrics scaled to 0--100. \textbf{Frm}: 
selected frames per video after Stage 2 filtering. \textbf{Time}: 
average end-to-end inference time per video in seconds. \textbf{Tok}: 
estimated visual tokens reaching the final summarization VLM, 
computed as (Frm $/$ 4 grouping)  using 
MiniCPM-V's average token cost across videos. Stage 2 incurs additional VLM captioning 
cost on the candidate frame set, not reported here.}
\label{tab:combined_ablation}

\end{table*}

We ablate PRISM on 25\% of YouCook2 and ActivityNet Captions, focusing on summary generation. To clarify the role of each component, we explicitly define the ablation settings. \textit{Video-Only} uses PRISM without transcript fusion and relies only on selected visual frames. \textit{No Stage 2} removes label generation and label-to-frame semantic anchoring, so summaries are generated from visually filtered frames selected by Stage 1. \textit{No Stage 1} removes ResNet- and PELT-based visual filtering and instead relies on semantic label-guided frame selection from all sampled frames (at 1 fps). \textit{No Processing} skips the frame-reduction pipeline and passes all frames directly to the summarization stage. Together, these settings isolate the effects of visual filtering, semantic anchoring, transcript fusion, and dense-frame processing on both summary quality and computational cost. All ablations use a Stage-2 label-confidence threshold of 0.9, a Stage-1 ResNet frame-difference threshold of 30, and an adaptive sampling batch size of 10. Results are reported in Table~\ref{tab:combined_ablation}, with all quality metrics scaled to 0--100. The time column reports the average end-to-end inference time per video, while the Tok column estimates the number of visual tokens passed to the final-stage summarization VLM. Full settings appear in Appendix~A, and hyperparameter analyses appear in Appendix~F.

\paragraph{Accuracy.}
The full PRISM pipeline (No Stage 2 setting, with final-stage 
grouping) matches or exceeds No Processing across all quality metrics 
on both datasets. On YouCook2, No Stage 2 achieves METEOR 31.83 and 
BERTScore 84.12, comparable to No Processing's 31.83 and 84.13 
despite using a fraction of the frames. On ActivityNet, No Stage 2 
achieves METEOR 17.59 and BERTScore 81.19, exceeding No Processing's 
16.09 and 80.56. We further find that Stage 1 alone is insufficient: 
since it relies only on motion cues, No Stage 1 outperforms No Stage 
2 on ROUGE-L, BERTScore, and METEOR for YouCook2, indicating that 
semantic label-guided selection contributes more to summary quality 
than motion-based filtering alone. The Video-Only modality 
outperforms many supervised baselines in 
Table~\ref{tab:youcook2_rouge_meteor} while using the fewest visual 
tokens, showing that label-guided selection retains semantic quality 
even under aggressive frame reduction. Its LLM-Judge score is lower 
on YouCook2 because the evaluator favors factual accuracy and 
complete step reconstruction, and many cooking actions and 
ingredient details are conveyed primarily in speech rather than 
visuals; adding transcripts (V+T) recovers this gap.

\paragraph{Efficiency.}
On YouCook2, the full pipeline reduces end-to-end inference time per 
video by 22.2$\times$ compared to No Processing (1{,}625.40s to 
73.09s) and selected frames by 81\% (322.91 to 60.07). Because the 
final stage groups frames into windows of four before summarization, 
visual tokens reaching the final-stage VLM drop from approximately 
207{,}000 to 9{,}600 per video, which is a 95.4\% reduction. Removing 
Stage 1 retains quality but inflates inference time substantially 
(236.46s vs.\ 73.09s), confirming that adaptive sampling drives the 
efficiency while semantic label validation in Stage 2 
provides the summary quality grounding.

On ActivityNet Captions, the full pipeline reduces inference time by 
4.6$\times$ (130.06s to 28.52s) and final-stage visual tokens from 
$\sim$72{,}000 to $\sim$3{,}200 per video, a 95.5\% reduction. 
ActivityNet contains short, motion-dense clips where motion patterns 
allow motion-based filtering alone to capture most meaningful 
events; the Video-Only modality also performs strongly here, using 
95.9\% fewer frames and 69\% less inference time, since visual 
information is informative when transcripts are sparse. This 
demonstrates that PRISM adapts to different video characteristics: 
on motion-driven datasets, motion-based filtering captures most 
events, while on procedurally complex datasets like YouCook2, the 
label-guided pathway and transcript fusion become essential.

\begin{figure}
    \centering
    \includegraphics[width=\columnwidth]{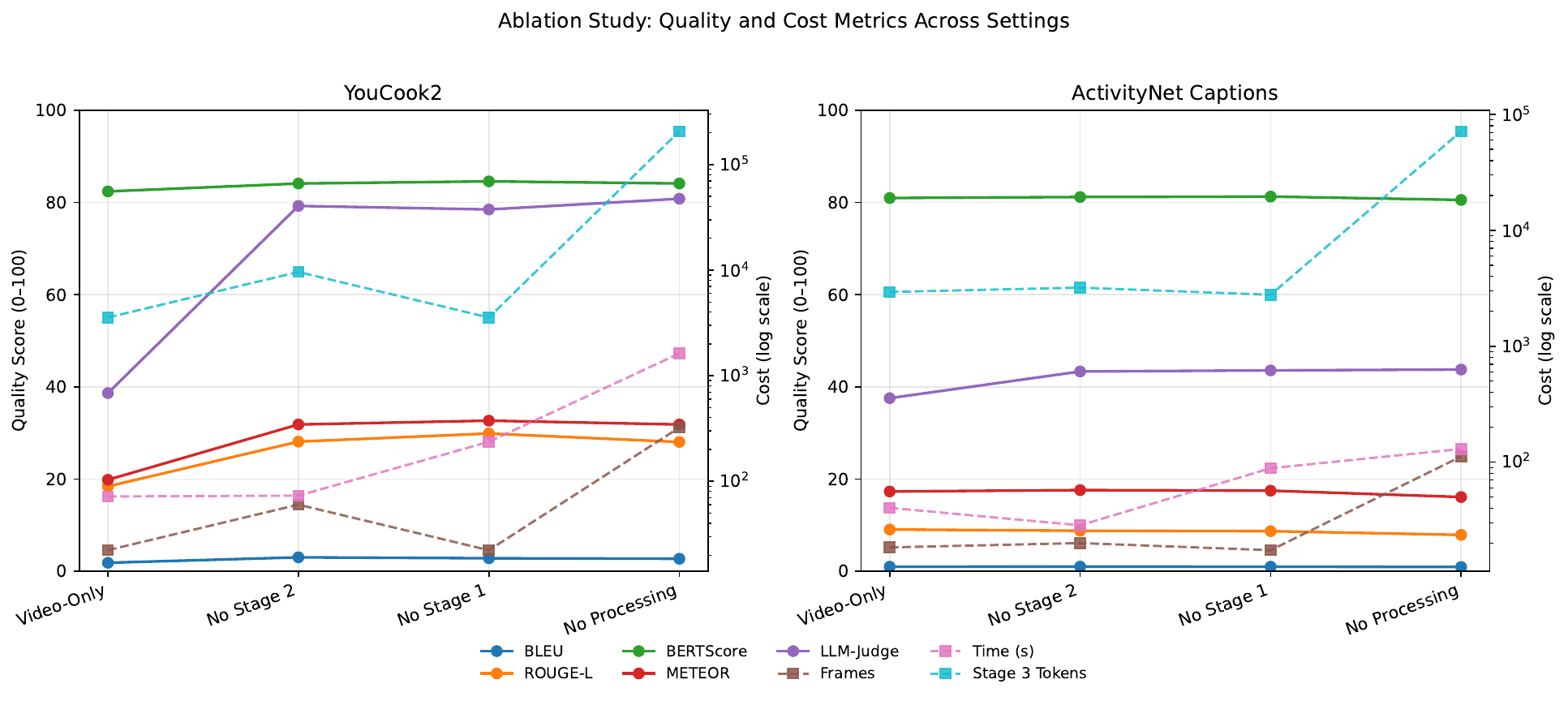}
    \caption{Ablation results on YouCook2 (left) and ActivityNet 
    (right). Solid lines: quality metrics on the left y-axis 
    (0--100). Dashed lines: cost metrics (frames, time, final-stage 
    visual tokens) on the right y-axis (log scale). Quality stays 
    stable across settings while cost rises by orders of magnitude 
    when frame selection is removed.}
    \label{fig:ablation_multiline}
\end{figure}

Figure~\ref{fig:ablation_multiline} visualizes this quality–cost 
trade-off across both datasets. Quality remains 
stable across the three label-guided settings; on YouCook2, 
METEOR varies by less than 1\% (31.83 to 32.66) and BERTScore by 
0.5\% across No Stage 2, No Stage 1, and No Processing, despite 
the two-order-of-magnitude difference in cost. Cost rises sharply when frame selection is removed, with final-stage 
visual tokens climbing from $\sim$10$^{3}$ to over $\sim$10$^{5}$ on 
YouCook2 under No Processing. The contrast between summary quality and exponentially rising cost illustrates PRISM's 
central trade-off: good frame reduction preserves summary 
quality, while the absence of frame reduction inflates compute cost. Final-stage grouping into windows of 
four is applied to all settings except No Processing.

\section{Conclusion}

We introduced PRISM, a training-free framework for frame-efficient procedural video summarization. PRISM combines lightweight visual filtering with dynamically generated procedural labels that act as semantic anchors for selecting informative frame-label pairs. Across our experiments, PRISM reduces visual inputs to fewer than 5\% of the original frames and lowers inference time by over 95\%, while preserving strong semantic content with BERTScore above 80\% and improving METEOR by up to 7.5\% over YouCook2 baselines. By moving frame reduction upstream of expensive VLM and language-model summarization stages, PRISM shows that efficient procedural video summarization can be achieved without supervised training, dataset annotations, or internal token compression. Future work will explore human evaluation of generated summaries and extensions to surgical report generation. 

\section{Limitations}

PRISM uses vision-language captioning and vision-encoder filtering, but consistency errors can still occur. In visually ambiguous settings, models may generate plausible but incorrect labels, such as mistaking ground beef for chicken, which can lead to a coherent but factually wrong summary. Dual-stage filtering reduces vague, redundant, and hallucinated outputs, but failures may remain in fine-grained or high-stakes domains. Domain-specific or fine-tuned VLMs could help mitigate these errors. 
\bibliography{cite}

\appendix
\section*{Appendix}

\subsection*{A. Experimental Settings}
All experiments were conducted on NVIDIA A100 GPUs (40GB HBM2 memory per GPU). The software stack consisted of Python 3.10, PyTorch 2.1, and CUDA 11.8. Inference of vision and language models was performed using the Ollama library and OpenAI GPT-4 API, with a temperature set to 0.0 for deterministic outputs where applicable. Data preprocessing and frame extraction used OpenCV, with all videos sampled at 1 frame per second (fps) for efficiency. ResNet18 was used with torchvision’s pre-trained models. 

\subsection*{B. Dataset Statistics}
Table~\ref{tab:dataset_simple_stats} presents key statistics for YouCook2 and ActivityNet Captions datasets, including the number of videos and average video duration. These statistics provide essential context for the scale and diversity of each dataset, helping to interpret experimental results and facilitate reproducibility. All ablation studies and sensitivity analyses in this work were performed exclusively on the validation sets, as test labels are not publicly available.

\begin{table}[h]
    \centering
    \setlength{\tabcolsep}{5pt}
    \begin{tabular}{lcc}
        \toprule
        \textbf{Dataset} & \textbf{\#Videos} & \textbf{Avg Length} \\
        \midrule
        YouCook2             & 414    & 5.26 \\
        ActivityNet Captions  & 4917  & 2 \\
        \bottomrule
    \end{tabular}
    \caption{Summary statistics for the official dataset. Avg Length denotes average time duration of the videos in minutes. \#Videos denotes number of videos. }
          \label{tab:dataset_simple_stats}

\end{table}
% 414 4917 

For all experiments, videos were sampled at 1 frame per second (fps) to ensure a balance between comprehensive temporal coverage and manageable computational resource usage. The combination of YouCook2 and ActivityNet Captions enables us to comprehensively evaluate our pipeline’s effectiveness in diverse real-world scenarios. YouCook2 consists of long-form cooking videos, where each video follows a detailed procedural sequence to fulfill a recipe. ActivityNet Captions encompasses a wide variety of activities, ranging from sports and hobbies to daily routines represented by much shorter video segments. This diversity allows us to demonstrate our pipeline’s ability to deliver robust content coverage and summarization quality across both long, structured instructional content and shorter, more varied event-driven videos, while operating under practical computational constraints. 

\subsection*{C. Evaluating Video Captioning Performance}

\begin{figure}[t]
    \centering
    \includegraphics[width=\columnwidth]{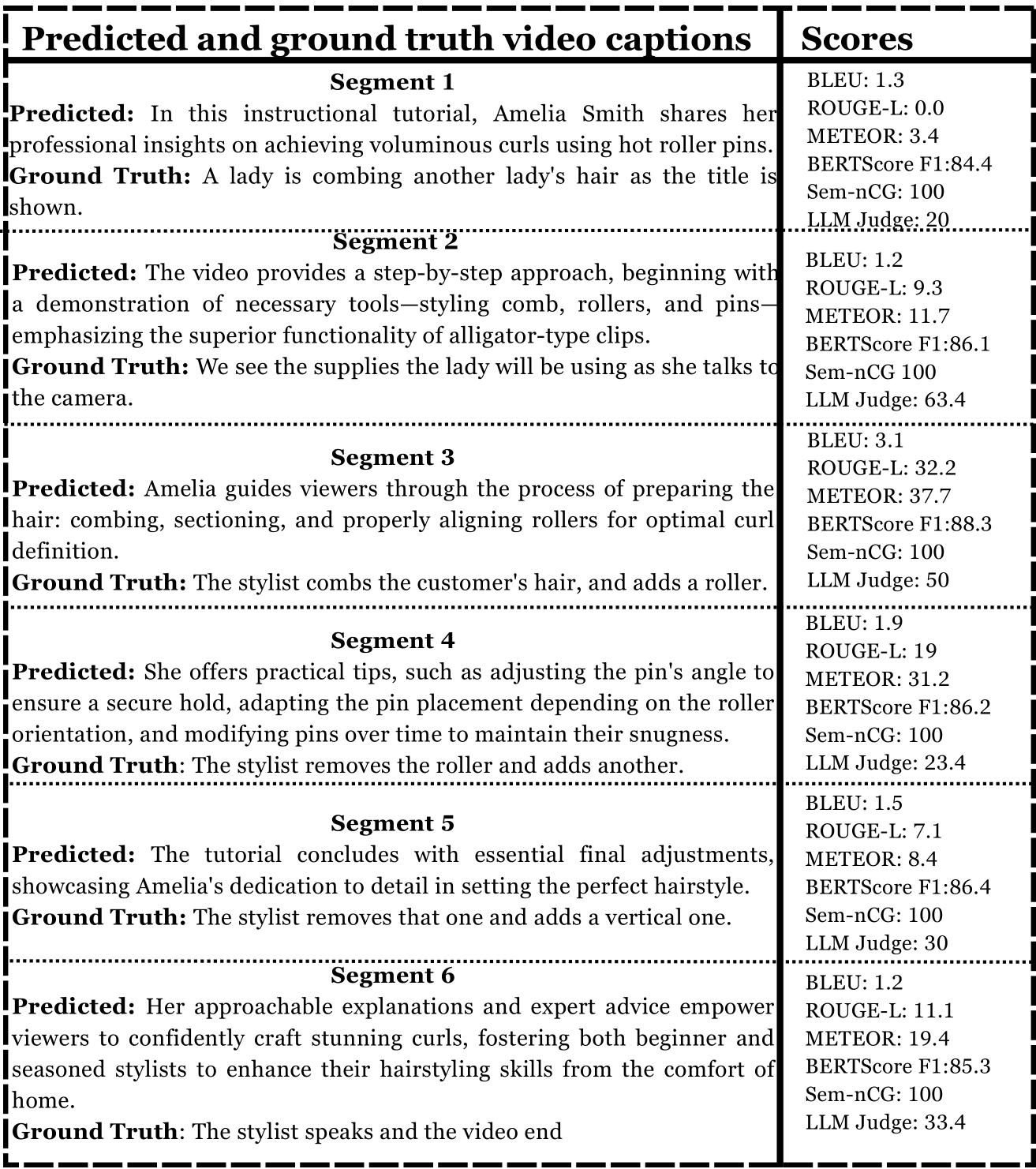}
    \caption{Predicted and ground-truth video captions with scores for a sample video from the ActivityNet Captions dataset.}
    \label{fig:activitynet_example}
\end{figure}

\begin{figure}[t]
    \centering
    \includegraphics[width=\columnwidth]{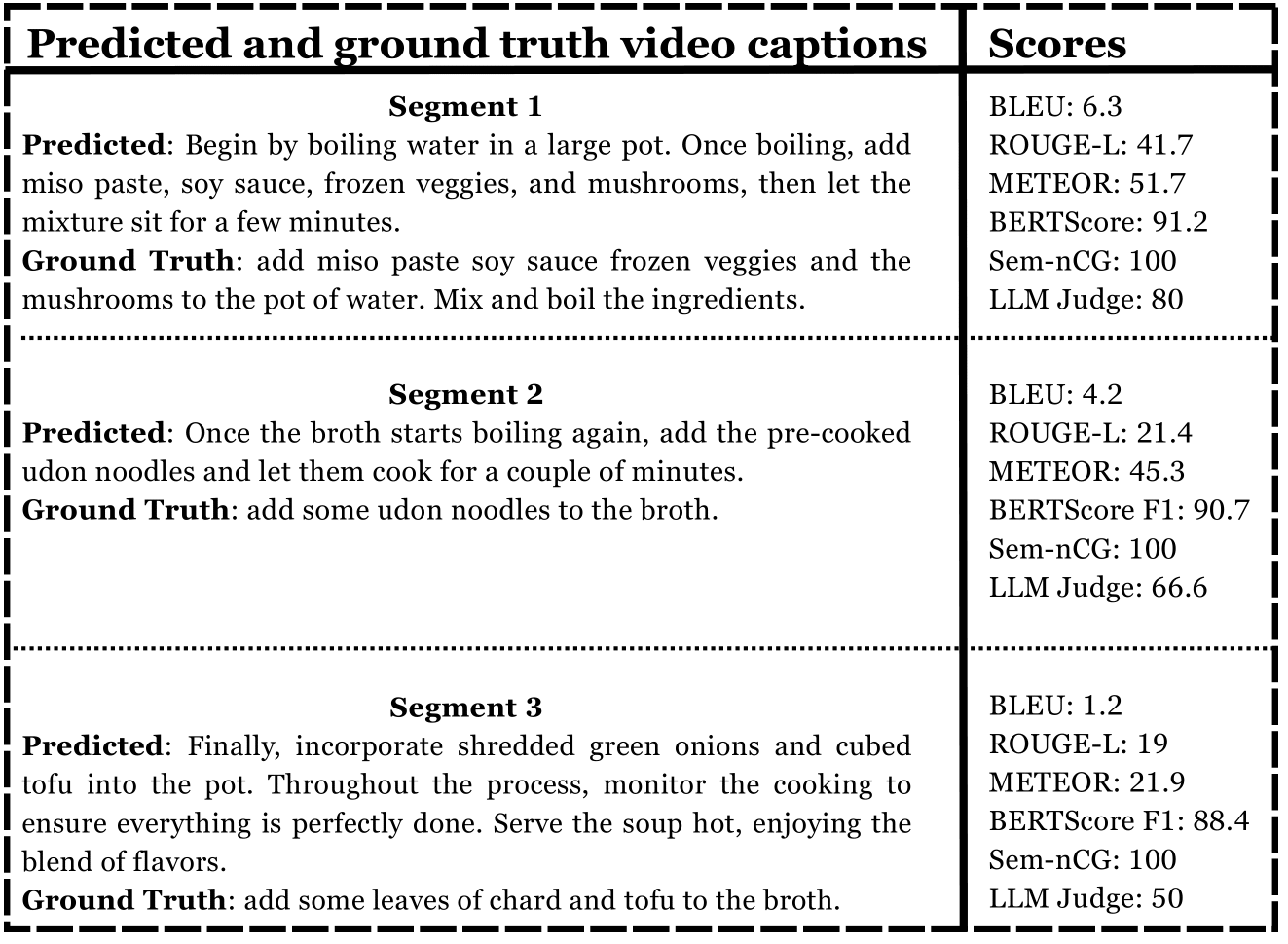}
    \caption{Predicted and ground-truth video captions with scores for a sample video from the YouCook2 dataset.}
    \label{fig:youcook2_example}
\end{figure}
All scores here are scaled to 0-100.
Figure~\ref{fig:activitynet_example} represents the PRISM-generated caption vs. ground truth on a sample video from ActivityNet Captions dataset; the predicted captions capture the instructional flow of the video, aligning well with the ground truth across segments. Despite minor lexical variations, high BERTScore and consistent LLM Judge scores indicate strong semantic overlap and overall quality. The model demonstrates a robust ability to generate coherent, contextually appropriate summaries.

Figure~\ref{fig:youcook2_example} shows the PRISM-generated caption vs. ground truth on a sample video from YouCook2 dataset. The predicted captions closely align with the ground truth, especially in capturing sequential steps, as reflected in the high BERTScore and LLM Judge values. This highlights the system’s effectiveness in both procedural understanding and summarization of activities or tasks with shorter durations.

\subsection*{D. Implementation Details}
For each dataset, videos were downloaded and processed according to the original dataset splits. Frames were extracted at 1 fps, then the frame feature difference threshold was set to 30 unless otherwise noted. Label assignment was conducted using the specified VLM (Gemma3, MiniCPM, GPT-4V), followed by GPT-4-based semantic filtering with temperature 0 for deterministic output. Label–frame pairs required cosine similarity $\geq 0.9$ with a vision encoder (e.g., CLIP) to be accepted. Sliding window-based summaries were generated and aggregated using LLMs as described in the main paper. No manual annotation or fine-tuning was performed.

\subsection*{E. Keyframe Selection}
Our pipeline targets video summarization with a focus on keyframe selection guided by semantic and temporal alignment. We evaluate this using the TVSum and SumMe datasets. TVSum consists of 50 user annotated videos across various categories with frame-level importance scores \citep{tvsum}, while SumMe contains 25 user-generated summaries for videos from various scenes of daily life \citep{SumMe}. 
Following the evaluation in the CSTA framework\citep{csta}, we align with their observation that traditional metrics like F1 score may not adequately capture the alignment between predicted and human-generated summaries, especially when ground truth annotations vary across users. Instead, we also adopt Kendall’s $\tau$ and Spearman’s $\rho$ to assess ranking correlation between predicted and ground truth importance scores.
Table~\ref{tab:summe_tvsum_comparison} presents a comprehensive comparison of the TVSum and SumMe datasets. These datasets contain short videos with diverse content and frame level human annotations. Our method, PRISM, ranks competitively among recent state-of-the-art models, particularly achieving $\tau = 0.1406$ and $\rho = 0.18$ on TVSum, while reaching $\tau = 0.1617$ and $\rho = 0.2212$ on SumMe. We attribute this to our algorithm’s ability to identify semantically rich anchors and perform label-aware clustering.
For the TVSum and SumMe experiments, we compute frame importance scores using a layered approach. First, we normalize ResNet feature differences. Then, during adaptive sampling, we re-normalize within frame groups. We also add CLIP-based vision encoder scores. At the grouping stage, we assign weights based on label distribution across four frames. If all labels are the same (1:1:1:1), each frame gets 0.25. For 3:1, the majority gets 0.75, the minority 0.25. In a 2:2 split, all receive 0.5. In 2:1:1, the majority gets 0.5, and others get 0.25.  All scores are averaged per frame, and if any frame gets dropped at a stage, it gets +0 for every stage after that.
\begin{table*}[t]
\centering
\scriptsize
\setlength{\tabcolsep}{2.2pt}
\renewcommand{\arraystretch}{0.78}

\begin{tabular}{lcccccc}
\toprule
\multirow{2}{*}{\textbf{Method}} 
& \multicolumn{3}{c}{\textbf{SumMe}} 
& \multicolumn{3}{c}{\textbf{TVSum}} \\
\cmidrule(lr){2-4} \cmidrule(lr){5-7}
& \textbf{Rank} & $\boldsymbol{\tau}$ & $\boldsymbol{\rho}$ 
& \textbf{Rank} & $\boldsymbol{\tau}$ & $\boldsymbol{\rho}$ \\
\midrule
Random & -- & 0.000 & 0.000 & -- & 0.000 & 0.000 \\
Human & -- & 0.205 & 0.213 & -- & 0.177 & 0.204 \\
dppLSTM~\citep{lstm} & 15 & 0.040 & 0.049 & 21 & 0.042 & 0.055 \\
HSA-RNN~\citep{hsarnn} & 12.5 & 0.064 & 0.066 & 19.5 & 0.082 & 0.088 \\
DAN~\citep{dan} & -- & -- & -- & 19.5 & 0.071 & 0.099 \\
DSNet-AB~\citep{dsn} & 14.5 & 0.051 & 0.059 & 15 & 0.108 & 0.129 \\
HMT~\citep{hmt} & 11.5 & 0.079 & 0.080 & 17.5 & 0.096 & 0.107 \\
CLIP-It~\citep{clipit} & -- & -- & -- & 13.5 & 0.108 & 0.147 \\
iPTNet~\citep{iptnet} & 9.5 & 0.101 & 0.119 & 11 & 0.134 & 0.163 \\
A2Summ~\citep{a2s} & 8 & 0.108 & 0.129 & 10.5 & 0.137 & 0.165 \\
VASNet~\citep{vasnet} & 7 & 0.160 & 0.170 & 9 & 0.160 & 0.170 \\
AAAM~\citep{aaam} & -- & -- & -- & 6.5 & 0.169 & 0.223 \\
MAAM~\citep{aaam} & -- & -- & -- & 5.5 & 0.179 & 0.236 \\
VSS-Net~\citep{vssnet} & -- & -- & -- & 3 & 0.190 & 0.249 \\
DMASum~\citep{dma} & 11 & 0.063 & 0.089 & \textbf{1} & \textbf{0.203} & \textbf{0.267} \\
RR-STG~\citep{rrstg} & 2.5 & 0.211 & 0.234 & 7.5 & 0.162 & 0.212 \\
MSVA~\citep{msva} & 3.5 & 0.200 & 0.230 & 5.5 & 0.190 & 0.210 \\
SSPVS~\citep{ssp} & 3$^\ast$ & 0.192 & 0.257 & 4.5 & 0.181 & 0.238 \\
GoogleNet~\citep{gnet} & 5 & 0.176 & 0.197 & 11.5 & 0.129 & 0.163 \\
CSTA~\citep{csta} & \textbf{1} & \textbf{0.246} & \textbf{0.274} & 2 & 0.194 & 0.255 \\
\midrule
\textbf{PRISM} & 6 & 0.162 & 0.221 & 9.5 & 0.141 & 0.180 \\
\bottomrule
\end{tabular}
\caption{Comparison with prior video summarization methods on SumMe and TVSum using rank, Kendall's $\tau$, and Spearman's $\rho$. T, M, and ST denote text-based, multimodal, and spatio-temporal methods, respectively.}
\label{tab:summe_tvsum_comparison}
\end{table*}
The abbreviations in brackets denote the input modalities used by each method: (T) indicates temporal features, (S) spatial features, and (M) multimodal inputs (e.g., vision-language or audio-visual). (ST) refers to models that jointly leverage both spatial and temporal features.

\subsection*{F. Hyperparameter Sensitivity}

We conduct hyperparameter sensitivity experiments using a stable configuration of Stage 2 threshold = 0.9, Stage 1 threshold = 30, and adaptive sampling batch size = 10, unless stated otherwise. When a specific setting is mentioned (e.g., Stage 2 = 0.5), only that parameter is varied, while the others remain fixed. These experiments are performed on both YouCook2 and ActivityNet Captions. All evaluation metrics and relevant frame-label statistics are reported to ensure transparency and reproducibility. All scores are scaled to 0-100.

We chose these defaults as a representative baseline, as a lower Stage 2 threshold like 0.5 might intuitively select more frames; in our setup the number of high-scoring frames remains stable due to the nature of the datasets and label distributions. In longer videos (e.g., 40 minutes or more) or high-stakes domains (e.g., surgical videos), strict thresholds (e.g., 0.9 in stage 2)  would likely filter out noise or ambiguity in frames, resulting in more significant differences.

For Stage 1, in the ResNet frame feature difference threshold, we compare thresholds of 10 vs. 30 vs. 50, where a lower value retains more candidate frames, and a higher value aims to significantly reduce redundancy. While experimenting with ResNet variants (ResNet-18, ResNet-50, etc.), we observed that deeper networks like ResNet-50 result in fewer frames passing Stage 1, reducing diversity in Stage 1 itself. Hence, we prioritize maintaining a diverse and sufficient sample for Stage 2 and downstream processing, settling on ResNet-18 as a practical trade-off. We also explore adaptive sampling batch sizes (10 vs. 15), where smaller sizes tend to generate more label diversity, and larger sizes produce more conservative label sets.

\begin{table*}[t]
\centering
\scriptsize
\setlength{\tabcolsep}{3pt}
\renewcommand{\arraystretch}{0.95}
\begin{tabular}{lcccccccc}
\toprule
\textbf{Setting} & \textbf{BERT} & \textbf{BLEU} & \textbf{METEOR} & \textbf{ROUGE} & \textbf{Sel.} & \textbf{Ext.} & \textbf{Filt.} & \textbf{Labels} \\
 & & & & & \textbf{Frames} & \textbf{Frames} & \textbf{Frames} & \textbf{/Video} \\
\midrule
Stage 2: 0.5 & 84.87 & 2.31 & 31.98 & 24.53 & 58.67 & 325.13 & 60.51 & 9.54 \\
Stage 2: 0.7 & 84.70 & 2.29 & 32.61 & 24.29 & 58.87 & 324.87 & 60.46 & 9.49 \\
Stage 2: 0.9 & 85.00 & 2.42 & 32.44 & 24.78 & 58.90 & 324.73 & 60.48 & 9.53 \\
Stage 1: 10  & \textbf{85.06} & 2.24 & \textbf{35.33} & \textbf{25.32} & 59.40 & 324.73 & 60.98 & 9.53 \\
Stage 1: 30  & 84.91 & 2.35 & 33.58 & 25.04 & 58.60 & 324.73 & 60.48 & 9.53 \\
Stage 1: 50  & 84.91 & 2.27 & 33.65 & 25.15 & 59.22 & 324.73 & 60.47 & 9.52 \\
Adaptive: 10 & 84.92 & 2.22 & 34.55 & 25.26 & 59.29 & 324.73 & 60.98 & 9.53 \\
Adaptive: 15 & 84.77 & 2.17 & 33.03 & 24.97 & 58.36 & 324.73 & 60.98 & 9.53 \\
\bottomrule
\end{tabular}
\caption{Hyperparameter sensitivity analysis on YouCook2. We report semantic and lexical metrics, selected/extracted/filtered frames, and labels per video.}
\label{tab:hyper_yc2}

\end{table*}
\begin{table*}[t]
\centering
\scriptsize
\setlength{\tabcolsep}{3pt}
\renewcommand{\arraystretch}{0.95}
\begin{tabular}{lcccccccc}
\toprule
\textbf{Setting} & \textbf{BERT} & \textbf{BLEU} & \textbf{METEOR} & \textbf{ROUGE} & \textbf{Sel.} & \textbf{Ext.} & \textbf{Filt.} & \textbf{Labels} \\
 & & & & & \textbf{Frames} & \textbf{Frames} & \textbf{Frames} & \textbf{/Video} \\
\midrule
Stage 2: 0.5 & 82.88 & 0.83 & 19.03 & 12.38 & 17.90 & 116.28 & 20.83 & 3.55 \\
Stage 2: 0.7 & 83.03 & 0.86 & 20.05 & 12.43 & 19.16 & 120.53 & 21.61 & 3.68 \\
Stage 2: 0.9 & \textbf{83.23} & \textbf{1.00} & 19.41 & 13.00 & 17.90 & 116.28 & 20.83 & 3.55 \\
Stage 1: 10  & 83.20 & 0.89 & 19.47 & 12.96 & 18.36 & 118.00 & 21.21 & 3.59 \\
Stage 1: 30  & 82.70 & 0.83 & 18.65 & 12.26 & 19.16 & 119.95 & 21.47 & 3.68 \\
Stage 1: 50  & 83.22 & 0.99 & 19.77 & 12.71 & 18.11 & 119.89 & 21.05 & 3.53 \\
Adaptive: 10 & 83.16 & 0.86 & 19.83 & 12.85 & 18.53 & 120.11 & 21.50 & 3.68 \\
Adaptive: 15 & 83.06 & 0.91 & \textbf{20.38} & \textbf{13.08} & 18.67 & 118.00 & 21.13 & 3.62 \\
\bottomrule
\end{tabular}
\caption{Hyperparameter sensitivity analysis on ActivityNet Captions. We report semantic and lexical metrics, selected/extracted/filtered frames, and labels per video.}
\label{tab:hyper_actnet}

\end{table*}

\vspace{1ex}
\noindent
\textbf{Note:} \textit{Stage 2} refers to the vision encoder similarity threshold , \textit{Stage 1} is the ResNet frame feature difference threshold, and \textit{adaptive} indicates the batch size for adaptive sampling.

\vspace{1em}

\noindent These tables provide a comprehensive view of how key hyperparameters affect both summary quality metrics and frame/label statistics on the two primary video summarization datasets in our study.
The hyperparameter sensitivity experiment shows that summary quality on YouCook2 and ActivityNet Captions remains consistent to changes in hyperparameters, with little effect on output quality. It should also be noted that in long-form videos (e.g., 40 minutes or more) with a higher visual similarity (e.g., colour consistency, less movement, etc.) amongst frames, small hyperparameter adjustments can result in significantly fewer or more selected frames. This highlights that while activities or instructional domains allow flexible tuning, careful hyperparameter selection is crucial for datasets with visually similar frames, especially in high-stakes domains.

\subsection*{G. Prompts used in the pipeline}
Throughout this code pipeline, we use prompts at multiple stages: beginning with frame-level visual descriptions via vision-language models, followed by label generation using GPT or LLaMA from those outputs, validation of those labels, detailed multi-frame image descriptions, and finally recursive and integrated summarization of chunked outputs to generate a coherent activity summary. The dataset description refers to a one-line description about the videos in the dataset (e.g., YouCook2: instructional cooking video).

\begin{itemize}
    \item \textbf{Frame Description Prompt}
    \begin{quote}
        Explain what is happening in the image. This is a frame from a video of an activity <dataset description>
    \end{quote}

    \item \textbf{Label Generation Prompt}
    \begin{quote}
        Extract info from: \{VLM\_output\}\\
        Tell me in few words (5–6) by giving a label for the most important details that you find from the text description of the activity happening in the image.\\
        Don't make the label vague like only the heading of the main activity; tell me exactly what is going on in the image and only give the label in simple words.
    \end{quote}

    \item \textbf{Label Validation Prompt}
    \begin{quote}
        Imagine you are an expert on validating labels. Given this label: \{label\}, do you think it is valid to check an image in an important video?\\
        For example, is it not useful (like a black screen) or too general (like ``this is a surgery video'' or ``this is a girl standing'')?\\
        I don't want labels that are not useful, say nothing about the image, or are too general.\\
        Based on your judgement, return 0 if you think this is unimportant or too general; return 1 if you think it is an important label. Return only a number, nothing else.
    \end{quote}
    
    \item \textbf{Combined Frame Description Prompt}
    \begin{quote}
        \textbf{Context:} This is a combination of 4 images from a video of an activity spanning <dataset description>\\
        Each small picture represents a step in the sequence:\\
        The current major label is '\{majority\_label\}'.\\
        - Top-right: Step 2\\
        - Top-left: Step 1\\
        - Bottom-right: Step 4\\
        - Bottom-left: Step 3\\
        Provide a detailed description for each of the 4 images.
    \end{quote}
    \item \textbf{Recursive Summarization Prompt}
    \begin{quote}
        You are summarizing partial text from a video of an activity spanning  <dataset description>.\\
        Try to give a name to this activity (e.g., a lady doing movements could be dancing). Try to correlate the different activities to speak for one main act or theme.\\
        Rewrite the text in a frame-wise, narrative format with transitions between steps:\\

        \{chunk\}
    \end{quote}

    \item \textbf{Final Integration Prompt}
    \begin{quote}
        We have a final summary of video frames:\\

        \{final\_summary\_text\}\\

        We also have the raw transcript from the entire audio:\\

        \{transcript\_text\}\\

        Please produce a unified, cohesive summary of all the steps in the activity happening in the video, which could be related to one of these: <dataset description>\\
        Incorporate relevant information from the audio transcript. If the transcript provides additional details or clarifications, weave them into the final summary.\\
        If the transcript includes extraneous content, omit it.\\
        Focus on a coherent storyline of the entire action or activity of the video.
    \end{quote}
\end{itemize}

\subsection*{H. Limitations and Error Analysis}

Despite the use of dual check mechanisms such as vision-language model-based captioning + vision encoder filtering, limitations still exist. One major challenge is model consistency, particularly with respect to hallucinations and mislabeling. Both vision-language and language models can generate plausible but incorrect outputs when operating without sufficient context. For instance, in a cooking video, the model may misinterpret ground beef as chicken based on visual similarity alone. Without audio cues or additional metadata, the generated summary may inaccurately describe the recipe as involving chicken, leading to a consistent but factually incorrect narrative.
These issues tend to surface in visually homogeneous or ambiguous settings, where subtle distinctions are difficult to detect. While our multimodal encoder-based filtering strategies help mitigate such errors, isolated failures still occur. In high-stakes domains, this can be mitigated to an extent using fine-tuned vision encoders and vision-language models. 
Additionally, the dual-stage validation process introduces increased computational overhead compared to simpler pipelines. However, this trade-off is necessary for improved reliability in the generated summaries.

\end{document}